\documentclass[conference]{IEEEtran}
\IEEEoverridecommandlockouts
\usepackage{cite}
\usepackage{amsmath,amssymb,amsfonts}
\usepackage{algorithm}
\usepackage{algpseudocode}
\algnewcommand\algorithmicforeach{\textbf{for each}}
\algdef{S}[FOR]{ForEach}[1]{\algorithmicforeach\ #1\ \algorithmicdo}
\usepackage{graphicx}
\usepackage{textcomp}
\usepackage{xcolor}
\usepackage{subcaption}
\usepackage{url}
\def\BibTeX{{\rm B\kern-.05em{\sc i\kern-.025em b}\kern-.08em
    T\kern-.1667em\lower.7ex\hbox{E}\kern-.125emX}}
\usepackage{cite}

\begin{document}

\title{
\vspace{-10mm}
{\normalsize \textcolor{red}{This abstract was accepted to and presented at the ``Multi-Agent Cooperative Systems and Swarm Robotics in the Era of Generative AI'' (MACRAI) workshop at the {\bf 2025 IEEE/RSJ Int. Conf. on Intelligent Robots and Systems (IROS 2025)}.\\ \vspace{5mm}}}
Online automatic code generation for robot swarms:\\ LLMs and self-organizing hierarchy\\
\thanks{Mary Katherine Heinrich and Marco Dorigo acknowledge support from the Belgian F.R.S.-FNRS, of which they are a Research Associate and a Research Director respectively.}
}

\author{\IEEEauthorblockN{Weixu Zhu}
\IEEEauthorblockA{IRIDIA\\
Université libre de Bruxelles\\
Brussels, Belgium\\
0000-0002-0329-9592}
\and
\IEEEauthorblockN{Marco Dorigo}
\IEEEauthorblockA{IRIDIA\\
Université libre de Bruxelles\\
Brussels, Belgium\\
0000-0002-3971-0507}
\and
\IEEEauthorblockN{Mary Katherine Heinrich}
\IEEEauthorblockA{IRIDIA\\
Université libre de Bruxelles\\
Brussels, Belgium\\
0000-0002-1595-8487}
}

\maketitle

\begin{abstract}
Our recently introduced {\it self-organizing nervous system (SoNS)} provides robot swarms with 1) ease of behavior design and 2) global estimation of the swarm configuration and its collective environment, facilitating the implementation of online automatic code generation for robot swarms. In a demonstration with 6 real robots and simulation trials with $>$30 robots, we show that when a SoNS-enhanced robot swarm gets stuck, it can automatically solicit and run code generated by an external LLM on the fly, completing its mission with an 85\% success rate.
\end{abstract}

\section{Introduction}

Swarm robotics research has demonstrated that many sophisticated behaviors with a large number of robots can be accomplished in a fully self-organized manner~\cite{dorigo2020reflections}, but these fully self-organized behaviors have been slow to transfer to real applications. One reason for this is the fact that robots in a swarm are programmed at the individual level but the desired behavior occurs at the group level, and the design of fully self-organized group behaviors is often analytically intractable~\cite{hamann2018swarm,brambilla2013swarm}, requiring extensive trial-and-error testing.

The long trial-and-error design process associated with self-organized behaviors is both a motivating factor and a complicating factor for automatic design. Because of the many testing trials typically required, automatic design for self-organized robot swarms is usually conducted offline~\cite{francesca2016automatic,hasselmann2021empirical}, and even progress made on online methods such as embodied evolution have often relied heavily on simulation~\cite{francesca2016automatic}. Besides the need for many trials, a second reason why offline methods remain common is the difficulty for a robot swarm to self-assess its own global performance. Robots in a self-organized swarm typically estimate or receive only their own individual states and the states of a few nearby robots. Using such local information, reaching a consensus on estimated global performance would require long convergence times impractical for most tasks. 

The inaccessibility of global information in a fully self-organized robot swarm also presents a substantial challenge for using large language models (LLMs) to generate code on the fly. Recent work has suggested, for example, using combinations of individual self-diagnosis and local peer-diagnosis for robots in a swarm to validate code generated by LLMs~\cite{strobel2024llm2swarm}. However, this does not resolve the difficulty of how to design individual behaviors that will result in the desired group behavior, without having access to global information.

In this paper, we propose that our recently introduced approach to self-organizing hierarchy---the {\it self-organizing nervous system (SoNS)}~\cite{zhu2024self}---can greatly simplify the implementation of online automatic code generation in robot swarms using LLMs. 
Using SoNS, robots can form and dissolve temporary centralized control structures in a self-organized manner~\cite{zhu2024self}.
Effectively, SoNS is a kind of middleware for robots to self-organize temporary, dynamic hierarchies with their peers, providing robot swarms with some useful functionalities: using SoNS, robots can coordinate their collective sensing, actuation, and decision-making activities in a temporarily centralized way, without sacrificing the scalability, flexibility, and fault tolerance benefits normally associated with self-organization.
Because the SoNS allows a whole robot swarm to be programmed as if it were a single robot with a reconfigurable body~\cite{zhu2024self}, requests to an LLM for new code can be much more straightforward than what would be required in robot swarms with flat (i.e., single-level and fully decentralized) system architectures.

\begin{figure*}[t!]
\centering
\begin{subfigure}[t]{.73\textwidth}
  \includegraphics[height=6.4cm]{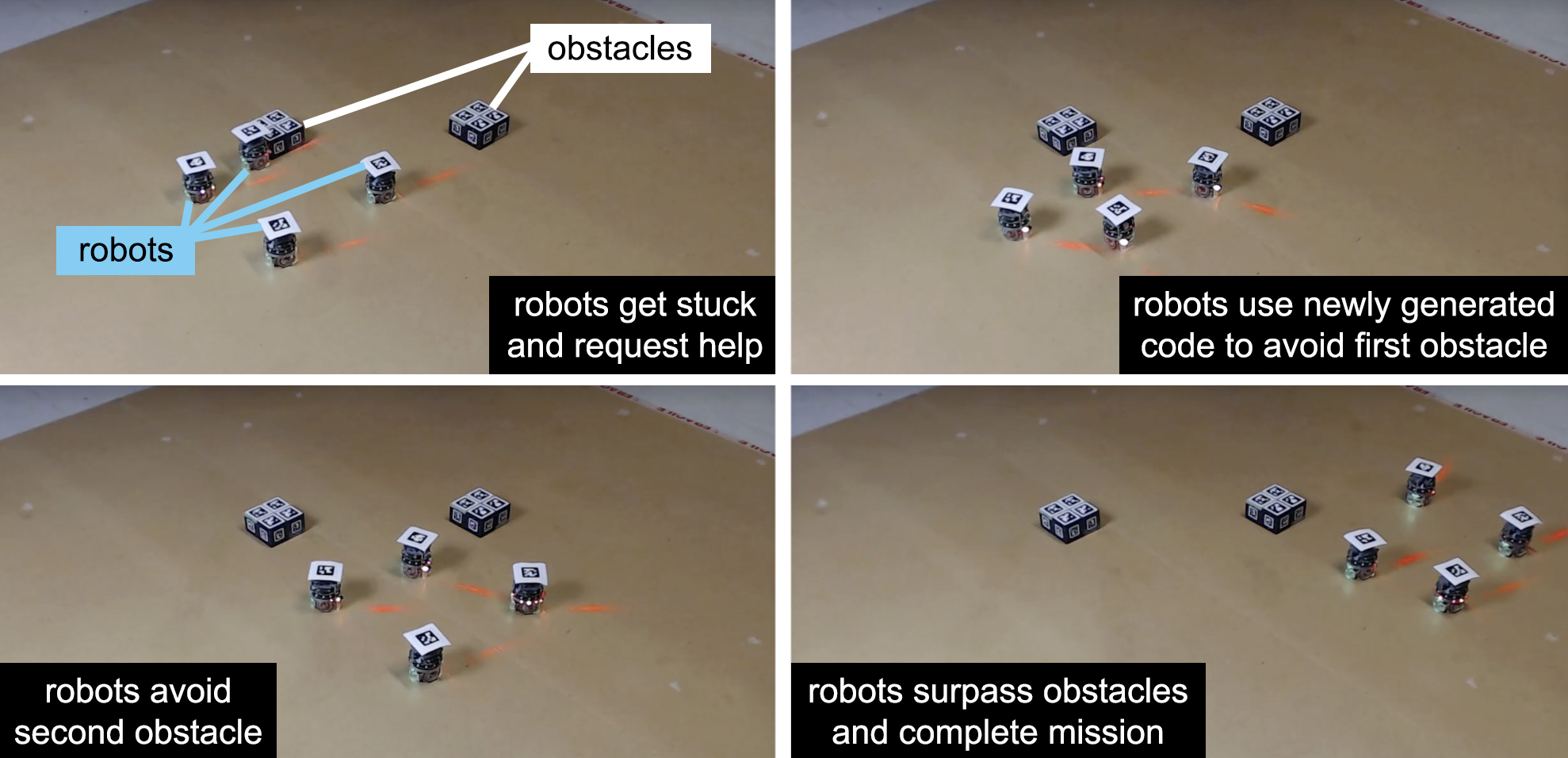}
  \caption{{\bf Demo with real robots.} Includes 4 ground robots, 2 tethered aerial robots [out of frame], and 2 obstacles. The real robots start the demo with the same software that the robots in a simulated trial start with, using the setup from~\cite{zhu2024self}.
}
  \label{fig:real-demo}
\end{subfigure}
\hspace{2mm}
\begin{subfigure}[t]{.21\textwidth}
  \includegraphics[height=6.4cm]{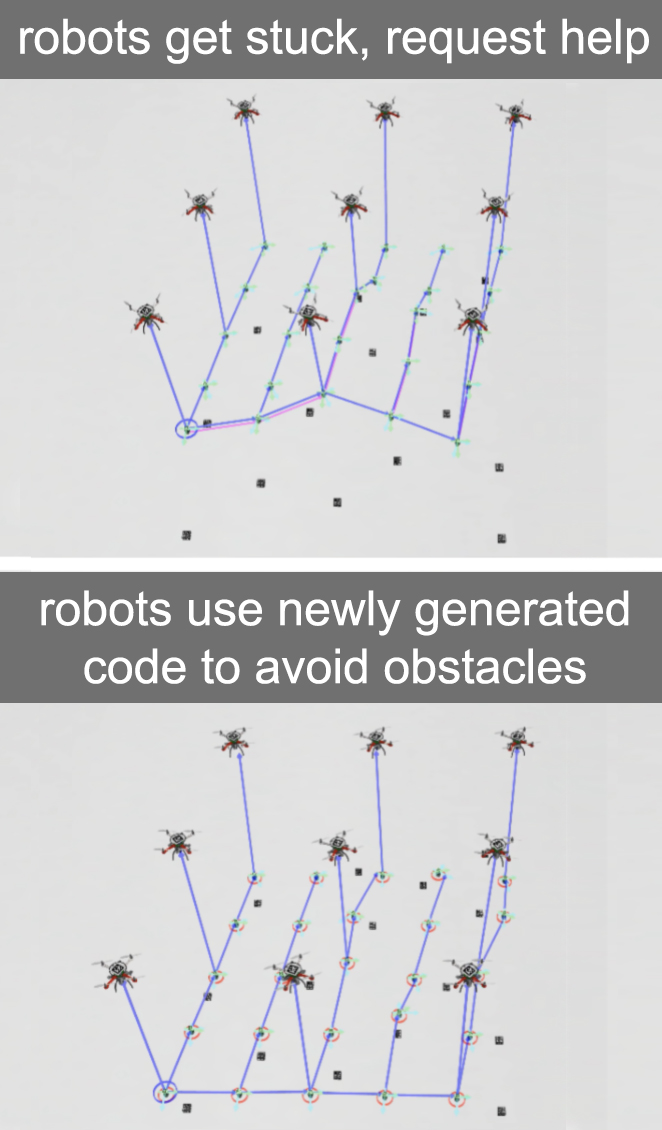}
  \caption{{\bf Simulation trial.} 25~ground robots, 9~aerial robots, 15 obstacles.}
  \label{fig:example-sim}
\end{subfigure}
\caption{\small{\bf Key frames from videos of the demo with real robots and an example successful simulation trial.} 
Robots begin with no code for obstacle avoidance; they simply move forward while remaining in a square formation shape. When the ground robots become physically obstructed, one robot sends its available information to an external LLM, with a generic request for new code for all robots in the swarm. Once the new code has been received and all robots in the swarm have updated their programs in a self-organized manner, the ground robots successfully circumvent the obstacles.}
\label{fig:keyframes}
\end{figure*}

\begin{figure}[h!]
\centering
{\includegraphics[width=0.46\textwidth]{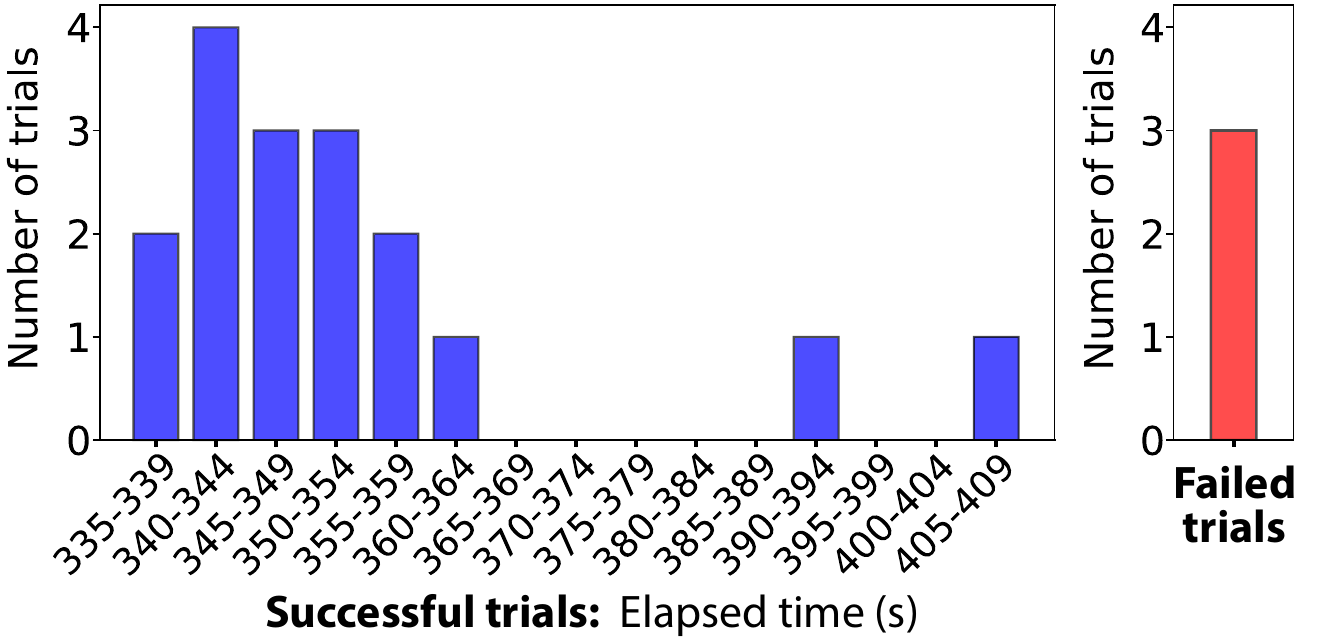}}
\caption{\small\textbf{Task duration using online LLM-based code generation, 20 trials.}  In successful trials, the elapsed time spans from when the first robot reaches the first obstacle, to when the last robot surpasses the last obstacle. Constituent steps include: 1) robots try and fail repeatedly to move forward; 2) robots send request for help to LLM; 3) robots receive and execute generated code; 4) robots get unstuck and surpass the obstacles, thus completing the task. Out of 20 trials, robots failed to complete the task in three (red bar).}
\label{fig:summary-sim}
\end{figure}

We propose that the SoNS~\cite{zhu2024self} is useful for online LLM-based generation of code for robot swarms in at least the following two ways:
\begin{enumerate}
    \item {\bf Ease of behavior design.} Because the self-organizing hierarchical structure of the SoNS separates global actuation from local actuation, an LLM can provide code for a desired global behavior directly, rather than attempting to construct a local behavior that when performed by many interacting robots will result in the emergence of the desired global behavior.
    \item {\bf Global estimation of swarm and environment.} Because sensor information is forwarded upstream in the self-organized network of a SoNS, culminating at an interchangeable brain robot, the brain robot can provide an LLM with an estimate of the global configuration of the whole swarm and its sensed environment (with a maximum update delay of $n+1$ steps, where $n$ is the depth of the rooted graph).
    \vspace{1mm}
\end{enumerate}

\section{Results and Discussion}

\begin{algorithm}[th!]
\footnotesize
\caption{\small\textbf{-- Methods: Program of robot $r_i$ for online code generation.} Functions for a robot to independently initiate a conversation with DeepSeek~R1 using the OpenRouter~API and request new code if the swarm gets stuck during operation. For definitions of SoNS variables, see~\cite{zhu2024self}.}
\label{alg:methods}
\begin{algorithmic}[1]
\vspace{1mm}
\Function{PrepareConversation}{}
\State {\bf declare} String as {\it ``I am a leader of a swarm, can you write some lua program to control the swarm for me?"}
\State {\bf append} description of the: SoNS context ($\boldsymbol{v}^{\textsc{Local}}$, $\boldsymbol{w}^{\textsc{Local}}$, $\boldsymbol{v}^{\textsc{Global}}$, $\boldsymbol{w}^{\textsc{Global}}$ velocity components), robot capabilities (differential drive, relative coordinate frame, sensing capabilities), known environment components (dimensions of the robots and potential obstacles, desired safety distance to an obstacle), mission goal (move forward at a desired speed), and code format (functions needed and example of overall format)
\EndFunction
%%%%%%%%%%%%%%%%%%%%%%%%%%%%%%%%%%%%%%%%%%%%%%%%%%%%%%%%%%%%
\vspace{3mm}
\Function{RequestHelp}{}
\State {\bf append} positional information of all obstacles and all robots detected by robot $r_i$ to String
\State {\bf append} {\it ``There seems to be something wrong, can you check what happened and improve my code?"} to String
\State {\bf send} String to OpenRouter API endpoint for LLMs
\EndFunction
%%%%%%%%%%%%%%%%%%%%%%%%%%%%%%%%%%%%%%%%%%%%%%%%%%%%%%%%%%%%
\vspace{3mm}
\Function{Main}{}
\State {\bf initiate} SoNS
\State \textsc{PrepareConversation}
\State {\bf loop} \textsc{Mission}
\Function{Mission}{}
\State Move forward at $|\boldsymbol{v}|$ target speed while maintaining formation
\If{robot $r_i$ is the current \textsc{SoNSbrain}}
    \If{robot $r_i$ has stopped moving forward} 
    \State \textsc{Timer} $\leftarrow$ \textsc{Timer} $+1$ \EndIf
    \If{\textsc{Timer} $> \theta_1$}
    \State \textsc{RequestHelp}
    \State wait for reply from the LLM
    \State extract generated code from the reply
    \State update \textsc{Mission} program with new code
    \State send updated \textsc{Mission} program to neighbor robots
    \EndIf
\ElsIf{a new \textsc{Mission} program is received from a neighbor robot}   
\State update \textsc{Mission} program with new code
\State send updated \textsc{Mission} program to neighbor robots
\EndIf
\EndFunction
\EndFunction
%%%%%%%%%%%%%%%%%%%%%%%%%%%%%%%%%%%%%%%%%%%%%%%%%%%%%%%%%%%%
\end{algorithmic}
\end{algorithm}

\begin{algorithm}[th!]
\footnotesize
\caption{\small\textbf{-- Results: Example code generated online.} Algorithms returned by DeepSeek~R1 during two example trials. Note that all replies received from DeepSeek~R1 during the trials also included a preceding qualitative description of the code. For definitions of SoNS variables, see~\cite{zhu2024self}.}
\label{alg:results}
\begin{algorithmic}[1]
\vspace{1mm}
\Statex {\hspace{-6mm} \small Example 1}
\hrulefill
%%%%%%%%%%%%%%%%%%%%%%%%%%%%%%%%%%%%%%%%%%%%%%%%%%%%%%%%%%%%
\vspace{2mm}
\footnotesize
\Function{Mission}{}
\ForEach{obstacle detected by robot $r_i$}
\If{detected distance to the obstacle is $<\theta_2$}
\State generate $\boldsymbol{v}^{\textsc{Local}}$, $\boldsymbol{w}^{\textsc{Local}}$ away from the obstacle
\If{$|0-x| < \theta_3$ for the $x$ component of $\boldsymbol{v}^{\textsc{Local}}$}
\State update $\boldsymbol{v}^{\textsc{Local}}$ to increase $|0-x|$ 
\State \Comment{due to using differential drive robots}
\EndIf \EndIf
\EndFor 
\If{robot $r_i$ is the current \textsc{SoNSbrain}}
\State \textsc{MoveForward}($\boldsymbol{v}^{\textsc{Global}}$, $\boldsymbol{w}^{\textsc{Global}}$)
\State send updated $\boldsymbol{v}^{\textsc{Global}}$, $\boldsymbol{w}^{\textsc{Global}}$ to children robots
\EndIf
\EndFunction
\vspace{1.5mm}
\vspace{2mm}
\Statex {\hspace{-6mm} \small Example 2}
\hrulefill
%%%%%%%%%%%%%%%%%%%%%%%%%%%%%%%%%%%%%%%%%%%%%%%%%%%%%%%%%%%%
\vspace{2mm}
\footnotesize
\Function{Mission}{}
\ForEach{obstacle detected by robot $r_i$}
\If{detected distance to the obstacle is $<\theta_2$}
\State generate $\boldsymbol{v}^{\textsc{Local}}$, $\boldsymbol{w}^{\textsc{Local}}$ away from the obstacle
\EndIf
\EndFor
\If{robot $r_i$ is the current \textsc{SoNSbrain}}
\If{no obstacles are in front within detected distance $\theta_3$}
\State \textsc{MoveForward}($\boldsymbol{v}^{\textsc{Global}}$, $\boldsymbol{w}^{\textsc{Global}}$)
\State send updated $\boldsymbol{v}^{\textsc{Global}}$, $\boldsymbol{w}^{\textsc{Global}}$ to children robots
\Else 
\State generate $\boldsymbol{v}^{\textsc{Global}}$, $\boldsymbol{w}^{\textsc{Global}}$ towards side with fewer obstacles
\State \textsc{MoveForward}($\boldsymbol{v}^{\textsc{Global}}$, $\boldsymbol{w}^{\textsc{Global}}$) at reduced speed
\State send updated $\boldsymbol{v}^{\textsc{Global}}$, $\boldsymbol{w}^{\textsc{Global}}$ to children robots
\EndIf \EndIf
\EndFunction
%%%%%%%%%%%%%%%%%%%%%%%%%%%%%%%%%%%%%%%%%%%%%%%%%%%%%%%%%%%%
%%%%%%%%%%%%%%%%%%%%%%%%%%%%%%%%%%%%%%%%%%%%%%%%%%%%%%%%%%%%
\end{algorithmic}
\end{algorithm}

In proof-of-concept demonstrations, we show that a SoNS-enhanced robot swarm can automatically solicit and run code generated online using a generic web API to an external LLM. The mission goal is simply to move forward while maintaining a square formation shape. At the start of a trial, the swarm has no code for obstacle avoidance and no {\it a priori} knowledge of the locations of obstacles. If the swarm gets stuck and cannot progress with its mission, the current SoNS-brain robot initiates a conversation with the LLM, sending the LLM all the basic information it has access to (about itself, the hierarchical organization of the swarm's actuation, the mission goal, and the environment, see Alg.~1 {\it Methods}), then sends all its current sensor information to the LLM along with a generic request for help. The SoNS-brain then updates its mission-specific program with the generated code returned by the LLM and sends the updated program to the robots it is directly connected to, so that the generated code will be spread to all robots in the swarm in a self-organized manner. The swarm then continues operation, and if it becomes stuck again, again requests help from the LLM, until the mission is complete. 

We demonstrate online LLM-based generation of code in heterogeneous aerial-ground robot swarms, as follows: 
\begin{itemize}
    \item {\bf 1 demonstration with 6 real robots}, 4 ground robots and 2 aerial robots in an environment with 2 unknown obstacles (see Fig.~\ref{fig:real-demo}); and
    \item {\bf 20 repetitions in simulation with 34 robots}, 25 ground robots and 9 aerial robots in an environment with 15 unknown obstacles (see Fig.~\ref{fig:example-sim}), with some variety in the algorithms returned by the LLM (see Alg.~2 {\it Results}) and an 85\% mission success rate (see Fig.~\ref{fig:summary-sim}).
\end{itemize}
We use the same robots and experiment setups as~\cite{zhu2024self}, except that in this paper the aerial robots are tethered to the ceiling in the real-robot demo; for details on the robots, experiment setup, and the SoNS approach and algorithms, please see~\cite{zhu2024self}. 
For details of the robots' program that are relevant to the LLM conversation, we provide the pseudocode (see Alg.~1 {\it Methods}).
For the external LLM, we use DeepSeek R1~\cite{guo2025deepseek} over the OpenRouter API\footnote{https://openrouter.ai/docs/quickstart} for interfacing with LLMs over a single endpoint, and the code requested from the LLM was specified to be in the Lua programming language. The results data (including videos of all trials and all LLM conversation logs) are available in an open-access online repository\footnote{\url{https://doi.org/10.5281/zenodo.17257762}}.

Although the robots provided context about their operation and their current sensor information to the LLM, the request they made to the LLM was fairly open-ended: {\it ``There seems to be something wrong, can you check what happened and improve my code?"} (see Alg.~1 {\it Methods}). In the 20 simulation trials, there was some variety in the code the LLM returned (see examples of two differing algorithms returned by the LLM in Alg.~2 {\it Results}), with some strategies enabling the robots to complete the task more quickly (see Fig.~\ref{fig:summary-sim}). In the three trials that we consider unsuccessful (see Fig.~\ref{fig:summary-sim}), the LLM returned code that removed mechanisms that enabled the current SoNS-brain to detect when other robots in the swarm were stuck, and thus some robots were left behind while the rest completed the mission. Future work could investigate more stringent separation of static code and update-able code, especially when the safety requirements are more complicated than the simple safety distances used in this paper. Also, in this paper, robots requested help anytime they got stuck and the behaviors the LLM returned were simple. Future work could study more principled approaches to requests for help and more complicated tasks.

\bibliographystyle{IEEEtran}
\bibliography{reference}

\begin{thebibliography}{1}
\providecommand{\url}[1]{#1}
\csname url@samestyle\endcsname
\providecommand{\newblock}{\relax}
\providecommand{\bibinfo}[2]{#2}
\providecommand{\BIBentrySTDinterwordspacing}{\spaceskip=0pt\relax}
\providecommand{\BIBentryALTinterwordstretchfactor}{4}
\providecommand{\BIBentryALTinterwordspacing}{\spaceskip=\fontdimen2\font plus
\BIBentryALTinterwordstretchfactor\fontdimen3\font minus \fontdimen4\font\relax}
\providecommand{\BIBforeignlanguage}[2]{{%
\expandafter\ifx\csname l@#1\endcsname\relax
\typeout{** WARNING: IEEEtran.bst: No hyphenation pattern has been}%
\typeout{** loaded for the language `#1'. Using the pattern for}%
\typeout{** the default language instead.}%
\else
\language=\csname l@#1\endcsname
\fi
#2}}
\providecommand{\BIBdecl}{\relax}
\BIBdecl

\bibitem{dorigo2020reflections}
M.~Dorigo, G.~Theraulaz, and V.~Trianni, ``Reflections on the future of swarm robotics,'' \emph{Science Robotics}, vol.~5, no.~49, 2020.

\bibitem{hamann2018swarm}
H.~Hamann, \emph{Swarm robotics: A formal approach}.\hskip 1em plus 0.5em minus 0.4em\relax Springer, 2018.

\bibitem{brambilla2013swarm}
M.~Brambilla, E.~Ferrante \emph{et~al.}, ``Swarm robotics: a review from the swarm engineering perspective,'' \emph{Swarm Intelligence}, vol.~7, 2013.

\bibitem{francesca2016automatic}
G.~Francesca and M.~Birattari, ``Automatic design of robot swarms: achievements and challenges,'' \emph{Front. in Robot. and AI}, vol.~3, 2016.

\bibitem{hasselmann2021empirical}
K.~Hasselmann, A.~Ligot \emph{et~al.}, ``Empirical assessment and comparison of neuro-evolutionary methods for the automatic off-line design of robot swarms,'' \emph{Nature communications}, vol.~12, no.~1, 2021.

\bibitem{strobel2024llm2swarm}
V.~Strobel \emph{et~al.}, ``{LLM2Swarm}: robot swarms that responsively reason, plan, and collaborate through {LLMs},'' \emph{preprint arXiv:2410.11387}, 2024.

\bibitem{zhu2024self}
W.~Zhu, S.~O{\u{g}}uz, M.~K. Heinrich \emph{et~al.}, ``Self-organizing nervous systems for robot swarms,'' \emph{Science Robotics}, vol.~9, no.~96, 2024.

\bibitem{guo2025deepseek}
D.~Guo \emph{et~al.}, ``{DeepSeek-R1} incentivizes reasoning in {LLMs} through reinforcement learning,'' \emph{Nature}, vol. 645, no. 8081, 2025.

\end{thebibliography}

\end{document}